\documentclass{sig-alternate}

\usepackage{subfigure}
\usepackage{booktabs}
\usepackage{xcolor}

\newcommand\eg{\emph{e.g.}}
\newcommand\ie{\emph{i.e.}}

\usepackage{algorithm}
\usepackage[noend]{algorithmic}
\usepackage{url}

\makeatletter
\def\@copyrightspace{\relax}
\makeatother

\begin{document}
%

\title{QPass: a Merit-based Evaluation of Soccer Passes}

%
%
%
%
%

\numberofauthors{2} 
\author{
%
\alignauthor
Laszlo Gyarmati\\
       \affaddr{Qatar Computing Research Institute, HBKU}\\
       \email{lgyarmati@qf.org.qa}
\alignauthor
Rade Stanojevic\\
\affaddr{Qatar Computing Research Institute, HBKU}\\
\email{rstanojevic@qf.org.qa}
}


\maketitle
\begin{abstract}
	Quantitative analysis of soccer players' passing ability focuses on descriptive statistics without considering the players' real contribution to the passing and ball possession strategy of their team. Which player is able to help the build-up of an attack, or to maintain the possession of the ball? We introduce a novel methodology called QPass to answer questions like these quantitatively. Based on the analysis of an entire season, we rank the players based on the intrinsic value of their passes using QPass. We derive an album of pass trajectories for different gaming styles. Our methodology reveals a quite counterintuitive paradigm: losing the ball possession could lead to better chances to win a game. 
\end{abstract}




\section{Introduction}

The connection of the players is a crucial ingredient of any team sports. This is utmost valid in case of soccer, where teams are having hundreds of passes per game to design their gameplay. Despite the fact that passes are having a significant role in the game, the state-of-the-art evaluation of the quality of passes is way to simplistic. In case of the media, the passing capability and quality of a player are measured using metrics such as the number of passes and their success ratio. In case of the soccer industry, this is supplemented by metrics related to creating scoring opportunities like number of assists, key passes, and forward passes.


These benchmarks are easy to be interpreted, however, they are missing the context of the passes. For example, a player may have 50 successful lateral passes at its own half of the pitch, delivering eye-catching statistics without real impact on the team's performance. Similarly, solely focusing on the number of assists a player is making neglects the build-up of the attack in which other players might have had more significant role.

Previous works that analyze soccer-related datasets focus on the game-related performance of players and teams. Vast majority of these methodologies concentrate on descriptive statistics that capture some part of the players' strategy or focus on analyzing the outcome of the strategies that teams apply~\cite{pena2012network,narizuka2013statistical,lucey2013assessing,gyarmati2014,gyarmati2015,Wang2015,luceyquality}. A quantification of scoring opportunities called expected goals has been proposed in the soccer analytics domain~\cite{expected_goals}, however, passing ability is lacking such a metric.


The state-of-the-art quantitative analysis of players' passing ability does not consider the players' real contribution to the passing and ball possession strategy of their team. Which player is delivering passes into critical parts of the field?
We introduce a novel methodology called QPass to answer questions like this quantitatively. Our main contributions are:
\begin{itemize}
\item First, we propose a scheme to quantify the value of having the ball at a particular part of the field. The methodology is team-specific, \ie, it takes into account the way a team organizes its defense and attack.
\item Second, we introduce a metric called QPass that derives the intrinsic value of passes building on the computed field values.
\item Third, we determine pass trajectories for diverse playing styles and the identity of players applying them. We reveal players with the highest and lowest quality of passes.
\item Finally, we show that making unsuccessful passes is beneficial in some cases and such leads to better chances to win a soccer game.
\end{itemize}

Applying QPass in practice can increase the comprehension of the core value of passes and thus the efforts players are making to help their teams. Player and opponent scouting is a prime area to apply the findings of our methodology in practice.

\section{Methodology}\label{sec:algorithm}
We next introduce our methodology called QPass to evaluate the intrinsic value of soccer passes. The method has three main steps, namely
\begin{itemize}
	\item first, we create a partitioning of the field that reflects the playing style of a team;
	\item second, we derive the value of having or not having the ball at a particular part of the field;
	\item third, we evaluate the merit of a pass based on the change in the field values owed to a pass.
\end{itemize}

Our methodology uses event-based datasets of soccer games, where major events like passes and shots of the games are annotated. The events are accompanied by the identity of the players and location information, among others. 
We add new, virtual passes to the dataset if the end coordinates of two consecutive passes are not identical.
Let $p=(x_s,y_s,f_s,x_e,y_e,f_e) \in P_i$ denote a pass (either original or virtual), where $x_s$ and $y_s$ are the coordinates of the start of the pass, while $f_s$ is the field value of the starting position. Analogously, $x_e,y_e,f_e$ denote the same in case of the end of the pass.

We apply the first two steps of the methodology iteratively to reduce the impact of the granularity of field partitioning on the field values.
We show the high-level steps of our method in Algorithm~\ref{alg:high_level}. As inputs, we use all the passes and shots of the teams we are analyzing. We determine the partitioning of the pitch and the field values team by team. Let $P_i$ and $S_i$ denote the passes and shots of team $i$, respectively.
We combine all the passes ($\hat{P}_i$) and shots ($\hat{S}_i$) made by an opponent against team $i$.
We start with $c_{max}$ number of clusters and smoothly decrease their number to $c_{min}$ (with steps of $c_{step}$). In each iteration, we use the events of the teams, the actual clustering, and the derived field values of the previous round.

\begin{algorithm}[tb]
	\scriptsize
\caption{Deriving field values and clusters of a team}
\label{alg:high_level}
\begin{algorithmic}
\STATE \textbf{Input}: $P$, set of passes; $S$, set of shots; $T$, set of teams
\STATE \textbf{Output}: $C$, clustering of the field; $F$, field values;
\STATE $c \leftarrow c_{max}$
\STATE $C \leftarrow \oslash, F \leftarrow \oslash$
\FORALL{$t_i \in T$}
	\STATE $P_i \leftarrow passes\_of\_team(i)$
	\STATE $\hat{P}_i \leftarrow passes\_of\_opponent(i)$
	\STATE $S_i \leftarrow shots\_of\_team(i)$
	\STATE $\hat{S}_i \leftarrow shots\_of\_opponent(i)$
	\WHILE{$c \geq c_{min}$}
		\STATE $C \leftarrow create\_clustering(c,P_i,\hat{P}_i,F_i,\hat{F}_i)$
		\STATE $F \leftarrow compute\_field\_values(C,P_i,\hat{P}_i,S_i,\hat{S}_i)$
		\STATE $c \leftarrow c - c_{step}$
	\ENDWHILE
\ENDFOR
\end{algorithmic}
\end{algorithm}

\subsection{Partitioning the soccer field}
In case of soccer analytics, the de facto method of aggregating and grouping events is to use a grid that divides the field into uniform cells (\eg, 5-by-5 meters). The drawback of this technique is that it is reluctant to consider the differences of teams' strategies. 
To overcome this, we use a team specific partitioning of the field; we outline the method in Algorithm~\ref{alg:clustering}.

First, we extract the location information of the start and end of all the passes of the teams 
along with the value of these positions computed in the previous iteration. We scale the values and then we use the mini batch K-means algorithm~\cite{sculley2010web} to create $c$ clusters. Once we have the clusterings, for each pass we determine to which cluster their locations are belonging to. 
If a team loses the ball, the other team gains possession at the end location of an unsuccessful pass. These coordinates have to be transformed as the events of both teams are coded based on their direction of attack.
We train two, separate classifier to determine the clusters ($l_e$) these transformed locations are belonging to. 

\begin{algorithm}[tb]
	\scriptsize
\caption{$create\_clustering$: Partitioning the field based on the events of a team}
\label{alg:clustering}
\begin{algorithmic}
\STATE $P_1 \leftarrow P_i, F_1 \leftarrow F_i$, $P_2 \leftarrow \hat{P}_i, F_2 \leftarrow \hat{F}_i$
\FORALL{$i \in {1,2}$}
	\STATE $P_i \leftarrow P_i \cup F_i$, $P_s \leftarrow {(x_s,y_s,f_s) \in P_i}$, $P_e \leftarrow {(x_e,y_e,f_e) \in P_i}$
	\STATE $P \leftarrow min\_max\_scaling(P_s \cup P_e)$
	\STATE $C_i \leftarrow mini\_batch\_kmeans(P,c)$
\ENDFOR
\FORALL{$i \in {1,2}$}
	\FORALL{$p=(x_s,y_s,f_s,x_e,y_e,f_e) \in P_i$}
		\STATE $c_s \leftarrow C((x_s,y_s,f_s))$, $c_e \leftarrow C((x_e,y_e,f_e))$
	\ENDFOR
	\STATE $NN_i \leftarrow nearest\_neighbors(P_i)$
\ENDFOR
\FORALL{$p=(x_s,y_s,f_s,x_e,y_e,f_e) \in P_1$}
	\STATE $l_e \leftarrow {NN}_2(mirror(x_e,y_e))$
\ENDFOR
\FORALL{$p=(x_s,y_s,f_s,x_e,y_e,f_e) \in P_2$}
	\STATE $l_e \leftarrow {NN}_1(mirror(x_e,y_e))$
\ENDFOR	
\end{algorithmic}
\end{algorithm}

\begin{figure}[tb]
\centering
\includegraphics[clip=true, trim=1cm 2.5cm 1cm 3cm,width=4cm]{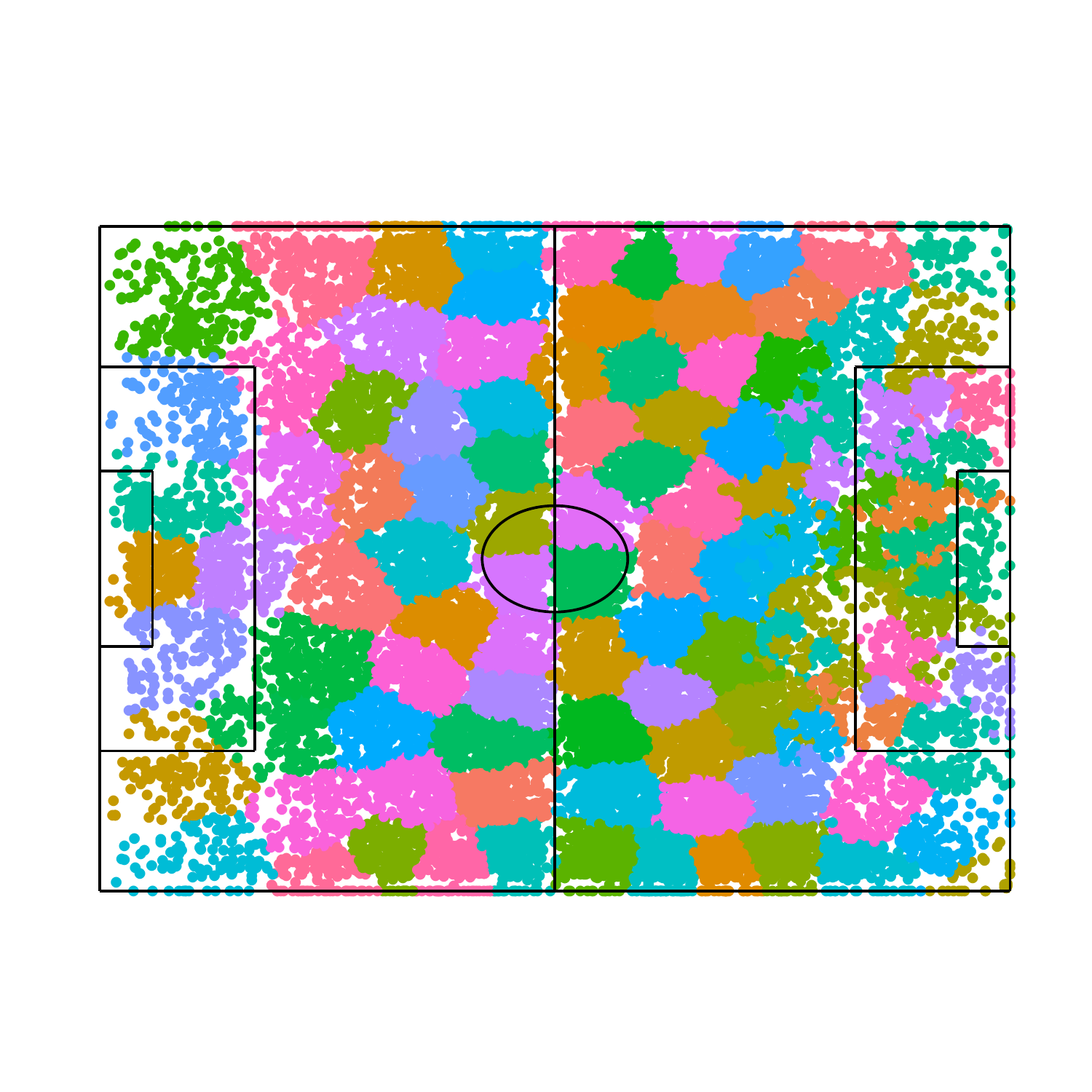}
\includegraphics[clip=true, trim=1cm 2.5cm 1cm 3cm,width=4cm]{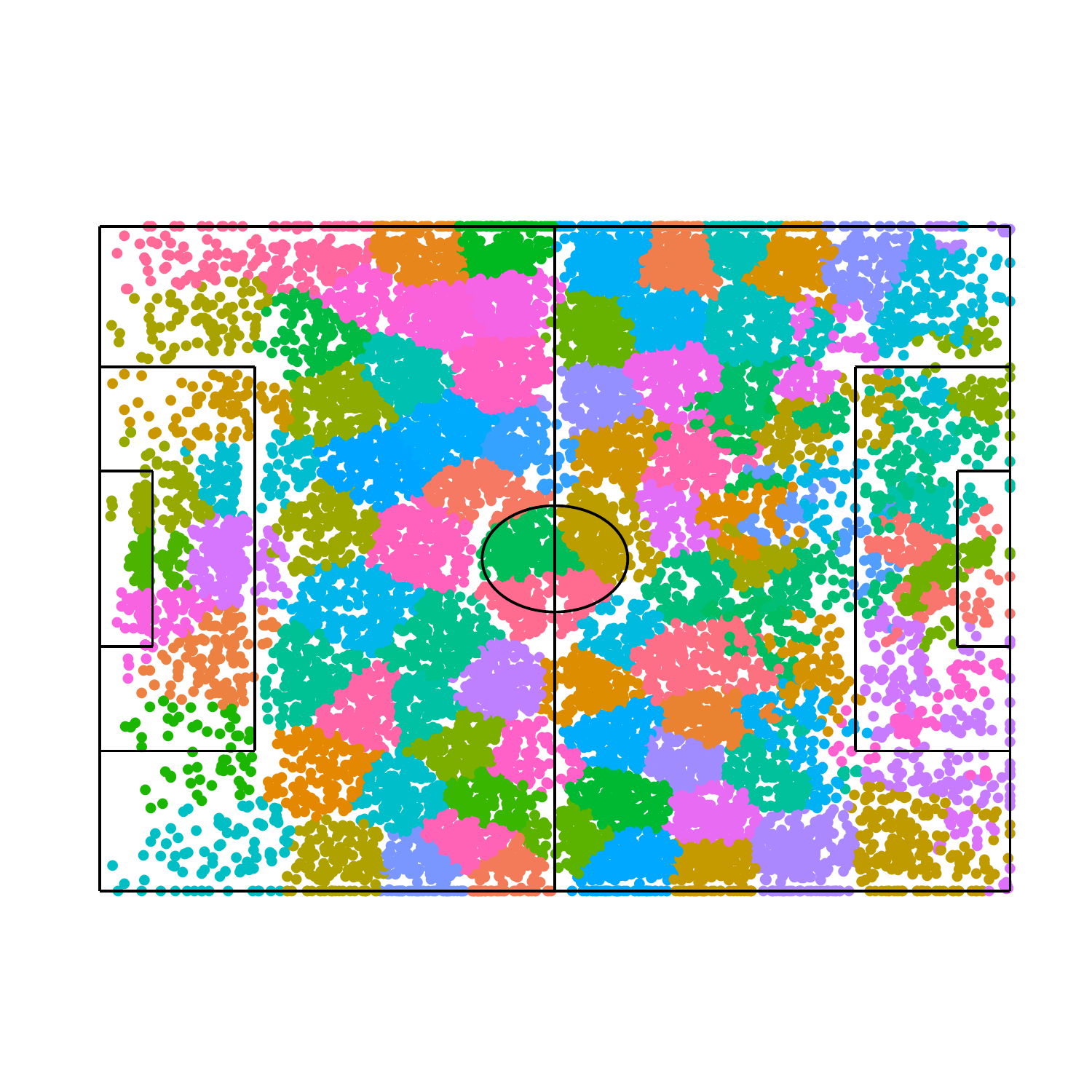}
\vspace{-5mm}
\caption{Team specific partitioning of the soccer field. The method considers the parts of the field teams are using more intensely. Teams attack from left to right. Left: FC Barcelona, right: Levante.}
\label{fig:clusters}
\end{figure}

In our evaluation, we use the following parameters to generate the clusters: $c_{max}=1000$, $c_{min}=100$, and $c_{step}=50$. We illustrate the partitioning of the field in case of two teams in Figure~\ref{fig:clusters}. FC Barcelona, the champion of the season, has more clusters and thus more fine-grained precision in the opponent's half of the pitch than Levante, the worst team of the league. 
For example, Barcelona has six clusters along the right sideline, two more than Levante, where the method focuses more on their own half of the pitch.

\subsection{Field value}
Using the partitioning, we compute the field values in case of two scenarios: a team has or does not have the ball at a specific part of the field. We compute the field values by solving a system of linear equations as shown in Algorithm~\ref{alg:field_values}. We use matrix $A$ to capture the transitions from one cluster to another, while vector $b$ stores the field values of the clusters. The first $c$ items of $b$ represent the field values when team $i$ has the ball, while the next $c$ items are for the cases where the opponents have the ball. Scoring or conceding a goal has a value of $1$ and $-1$. Similarly, let $s$ denote the value of taking a shot.
 
If a pass is successful, a transition happens from the cluster of the start of the pass to the cluster of the end of the pass. If the pass is the last one in a given ball possession (\ie, sequence of uninterrupted, consecutive passes of a team), we add an additional state transition describing the change of ball possession ($l_e + c$).
We also handle the shots based on their outcome. If a shot had an assist, we create a new transition from the end of the assist to the location of the shot.
The procedure is identical in case of the opponent team except the addressing of the appropriate states in the transitions. 
After considering all the potential state transitions, we normalize the transition matrix and finally solve the system of linear equations.

\begin{algorithm}[tb]
	\scriptsize
\caption{$compute\_field\_values$: Computing the field values given a clustering of the field}
\label{alg:field_values}
\begin{algorithmic}
\STATE $c \leftarrow |C|$, $A \leftarrow zeros(2c+4,2c+4)$, $b \leftarrow zeros(2c+4)$
\STATE $b_{2c+1} \leftarrow 1$, $b_{2c+2} \leftarrow s$, $b_{2c+3} \leftarrow -1$, $b_{2c+4} \leftarrow -s$
\FORALL{$i \in {1,2}$}
	\FORALL{$p=(c_s,c_e,l_e) \in \hat{P}_i$}
		\IF{$successful(p)$}
			\STATE $A(c_s,c_e) \leftarrow A(c_s,c_e) + 1$
			\IF{$last\_pass(p)$ \AND $no\_shot(p)$}
				\STATE $A(c_e,l_e) \leftarrow A(c_e,l_e) + 1$
			\ENDIF
		\ELSE
			\STATE $A(c_s,l_e) \leftarrow A(c_s,l_e) + 1$
		\ENDIF
	\ENDFOR
	\FORALL{$s=(c_s) \in \hat{S}_i$}
		\IF{$goal(s)$}
			\STATE $A(c_s,2c+1) \leftarrow A(c_s,2c+1) + 1$
		\ELSE 
			\STATE $A(c_s,2c+2) \leftarrow A(c_s,2c+2) + 1$
		\ENDIF
		\IF{$p=has\_assist(s)$}
			\STATE $A(c_e(p),c_s) \leftarrow A(c_e(p),c_s) + 1$
		\ENDIF
	\ENDFOR
\ENDFOR
\STATE $normalize(A)$
\STATE $solve(A,b)$
\end{algorithmic}
\end{algorithm}

The used system of linear equations has a nice feedback mechanism to determine the field value of the different clusters. The field value of a cluster depends on the field values of the clusters to where passes were made, on the number of shots and goals created, and on the field values of clusters where ball possessions were lost. However, the field value of not having the ball at a given cluster depends on the field values of multiple scenarios: both with and without ball possessions.

\begin{figure*}[tb]
\centering
\subfigure[Ball possession of FC Barcelona]{\includegraphics[clip=true, trim=1cm 2.5cm 1cm 3cm,width=5cm]{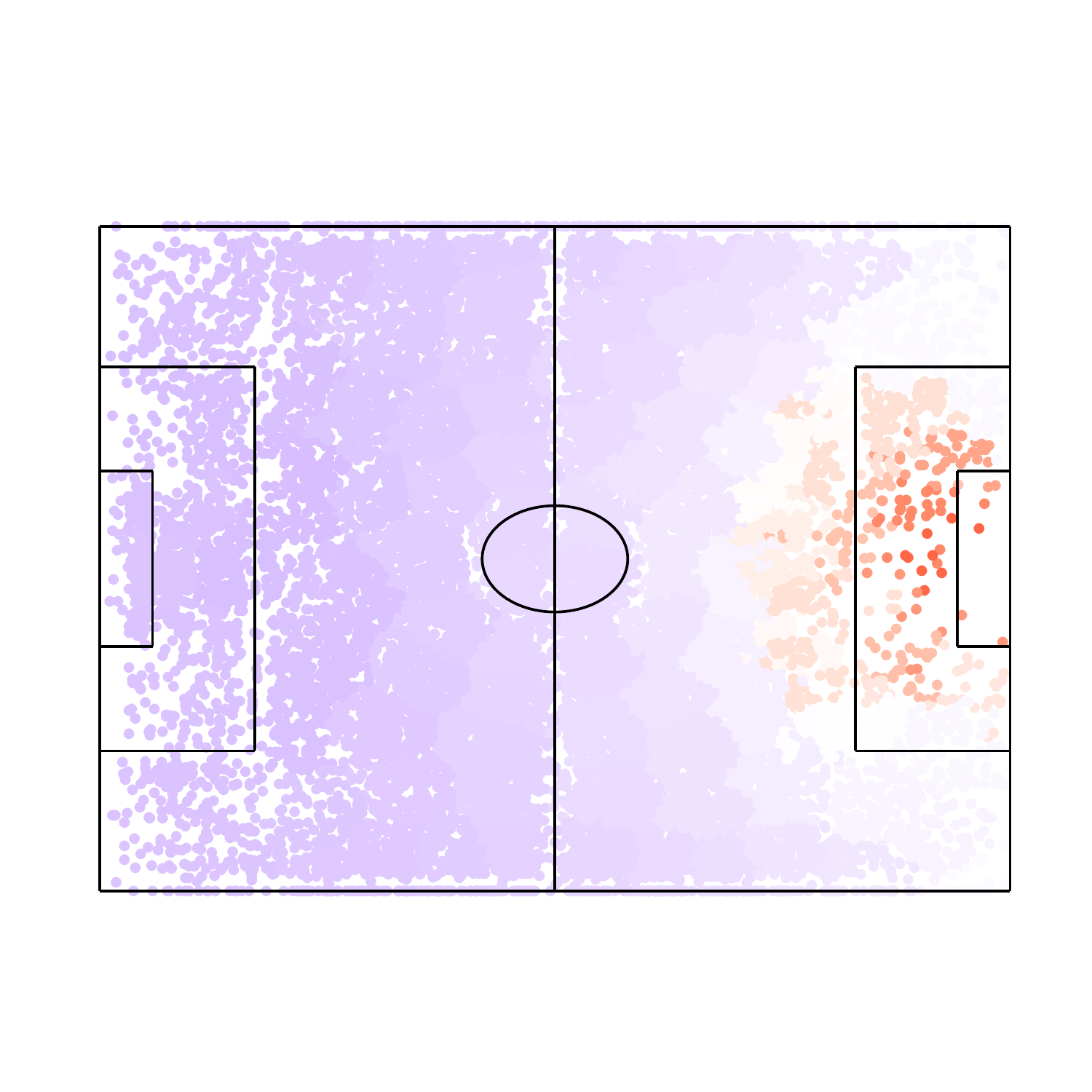}\label{fig:field_values_fcb}}
\subfigure[Ball at theopponents of FCB]{\includegraphics[clip=true, trim=1cm 2.5cm 1cm 3cm,width=5cm]{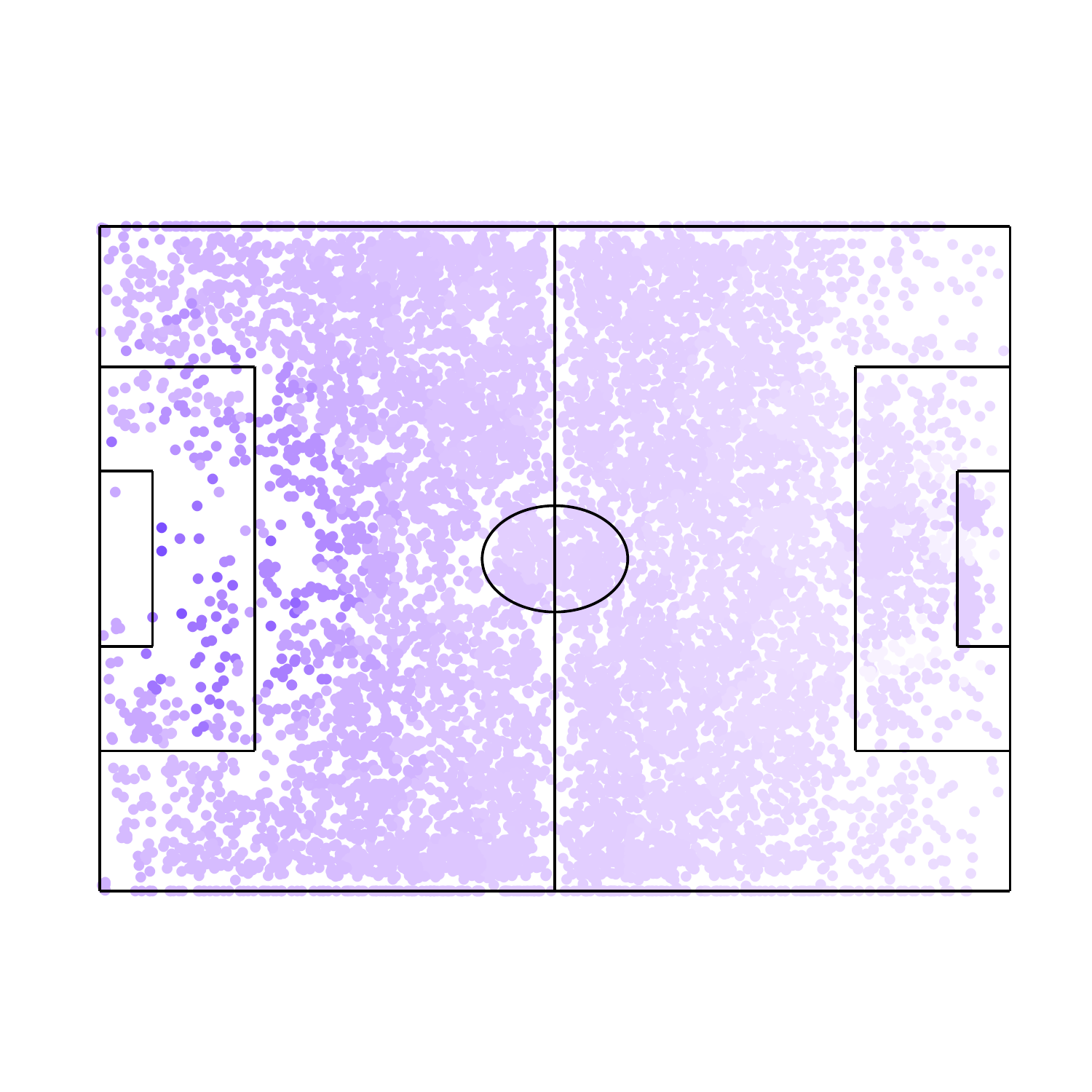}\label{fig:field_values_fcb_not}}
\subfigure[Ball possession of Levante]{\includegraphics[clip=true, trim=1cm 2.5cm 1cm 3cm,width=5cm]{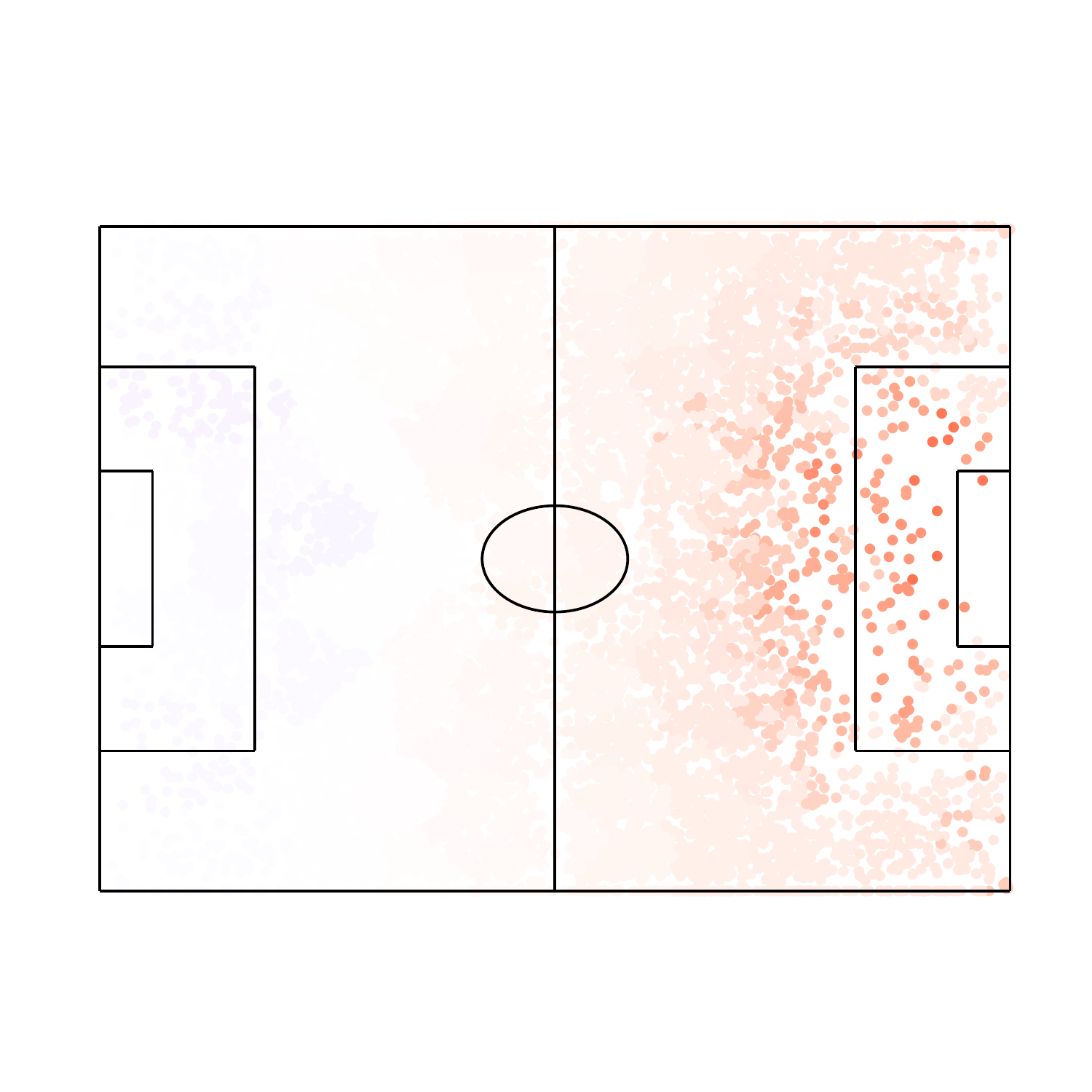}\label{fig:field_values_levante}}
\vspace{-5mm}
\caption{The values of parts of the field reflect both the attacking and the defending strategies of the teams. The intensity of the colors denote how beneficial the given location is; red represents the most worthy positions while blue areas are the least beneficial ones.}
\label{fig:field_values}
\end{figure*}

To illustrate the behavior of the field value metric, we present three scenarios in Figure~\ref{fig:field_values}. The ball possession of FC Barcelona is not that valuable up until the final 30 meters of the field, \ie, the field values are less then zero (Figure~\ref{fig:field_values_fcb}). The numerous passes FC Barcelona is making and their ball possession focused style (\ie, they are seldom having attacks with only few passes) cause this behavior. The figure reveals an imbalance between the two sides of their attacks: the team is better off having the ball on the right side of the field. Figure~\ref{fig:field_values_fcb_not} shows the field values of FC Barcelona, if they are not having ball. The behavior of the metric is inline with common sense: the farther the opponent controls the ball from the goal of FC Barcelona the better. 
We also show the field values of Levante in Figure~\ref{fig:field_values_levante}. The field values are positive if Levante has the ball at the opponent's half of the pitch---a harsh contrast to FC Barcelona. Teams like Levante have possession of the ball less frequently in these parts of the field and they usually apply direct attacks. 

\subsection{The value of a pass: QPass}
We use the derived field values to evaluate the merit of every pass. We compute QPass as the change in the field values as the result of a pass:
\begin{equation}
	\textrm{QPass}(p) = \left\{
	\begin{array}{rl}
	 f_e - f_s & \mbox{ if p is successful} \\
	 l_e - f_s & \mbox{ otherwise}
	 \end{array} \right.
\end{equation}
\noindent QPass depends not only on the location of the endpoints of a pass but also on the outcome of the pass. If the pass is unsuccessful, we use the field value of the end location given the opponent is having the ball possession.


\section{Empirical Results}
We next present an empirical evaluation of the QPass metric. We use an event-based dataset covering the 2015/16 season of the Spanish La Liga. The dataset contains all major events of a soccer game including passes and shots. The dataset has more than 330,000 passes and nearly 8,600 shots. 
We pick $s=0.7$ to be the value of having or conceding a shot. 
We only consider players who had at least 100 passes throughout the season. We use the median QPass values in our analysis because outlier values may exist due to the cluster driven computation of QPass. 

\begin{figure}[tb]
\centering
\includegraphics[width=7cm]{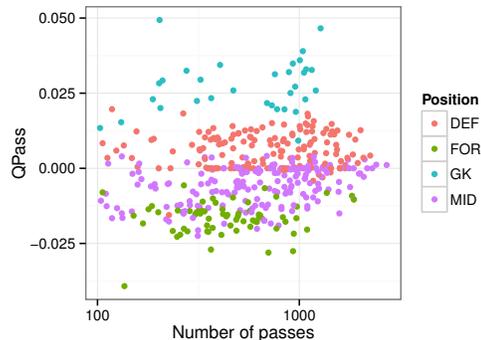}
\vspace{-6mm}
\caption{Median QPass values as a function of the number of passes players have throughout a season. There are clear separations among the positions of the players.}
\label{fig:qpass_median}
\end{figure}

First, we present the distribution of median QPass values based on the number of passes the players have (Figure~\ref{fig:qpass_median}). The number of passes has negligible impact on QPass values. The players' position however affects the median values: goalkeepers have the largest QPass while attackers usually cannot increase the field values via passing. Midfielders have positive or negative impact on field values.

\subsection{Ranking of players}
QPass is able to capture the diversity of the roles players have in a soccer team. We next focus on the best and worst players of each role with respect to the players' contribution to the play of the teams. Goalkeepers and defenders start the build-up of the attacks and also responsible to prevent the opponent's scoring chances. If a team plays in a direct style, players with high QPass are preferred such as T. Kolodziejczak (Sevilla, defender) or G. Iraizoz (Athletic Club, goalkeeper). 
For keeping the ball safe (\ie, having passes with low QPass values), midfielders like N. Amrabat of Malaga are preferred. 
The other end of the spectrum is S. Cristoforo (Sevilla), who is able to make passes into more beneficial parts of the field.

We show the case of attackers in Table~\ref{tab:ranking_attackers}. Attackers with the highest QPass values are crucial for creating chances. The ranking of QPass values reveal players who are capable and willing to create key attacking passes like L. Vazquez, A. Griezmann, or L. Messi. Players whose main role is to score are having the lowest median QPass values including R. Marti, A. Aduriz, and L. Suarez.

\begin{table}[tb]
	\caption{Top and bottom five attackers. 
	High QPass reveals the players' involvement in creating chances. Strikers have low QPass values: they are already at a prime location to score, an additional pass moves the ball to a less valuable part of the field.}
	\label{tab:ranking_attackers}
	\scriptsize
	\centering
\begin{tabular}{lrl}
\toprule
Player &  QPass (median) & Team  \\
\midrule
 Rodrigo de Paul & -0.005673  &         Valencia  \\
   Lucas Vazquez & -0.006612  &         Real Madrid  \\
   David Barral & -0.008006  &          Granada  \\
 Antoine Griezmann & -0.008176  &  Atletico Madrid  \\
  Lionel Messi & -0.009155  &           Barcelona  \\
\midrule                                 
    Nabil Ghilas & -0.022783  &        Levante  \\
 Fernando Llorente & -0.027046  &        Sevilla  \\
     Luis Suarez & -0.027562  &      Barcelona  \\
    Aritz Aduriz & -0.028058  &  Athletic Club  \\
    Roger Marti & -0.039211  &        Levante  \\
\bottomrule
\end{tabular}
\end{table}

\begin{figure}[tb]
\centering
\subfigure[Lionel Messi]{\includegraphics[clip=true, trim=1cm 2.5cm 1cm 3cm,width=4cm]{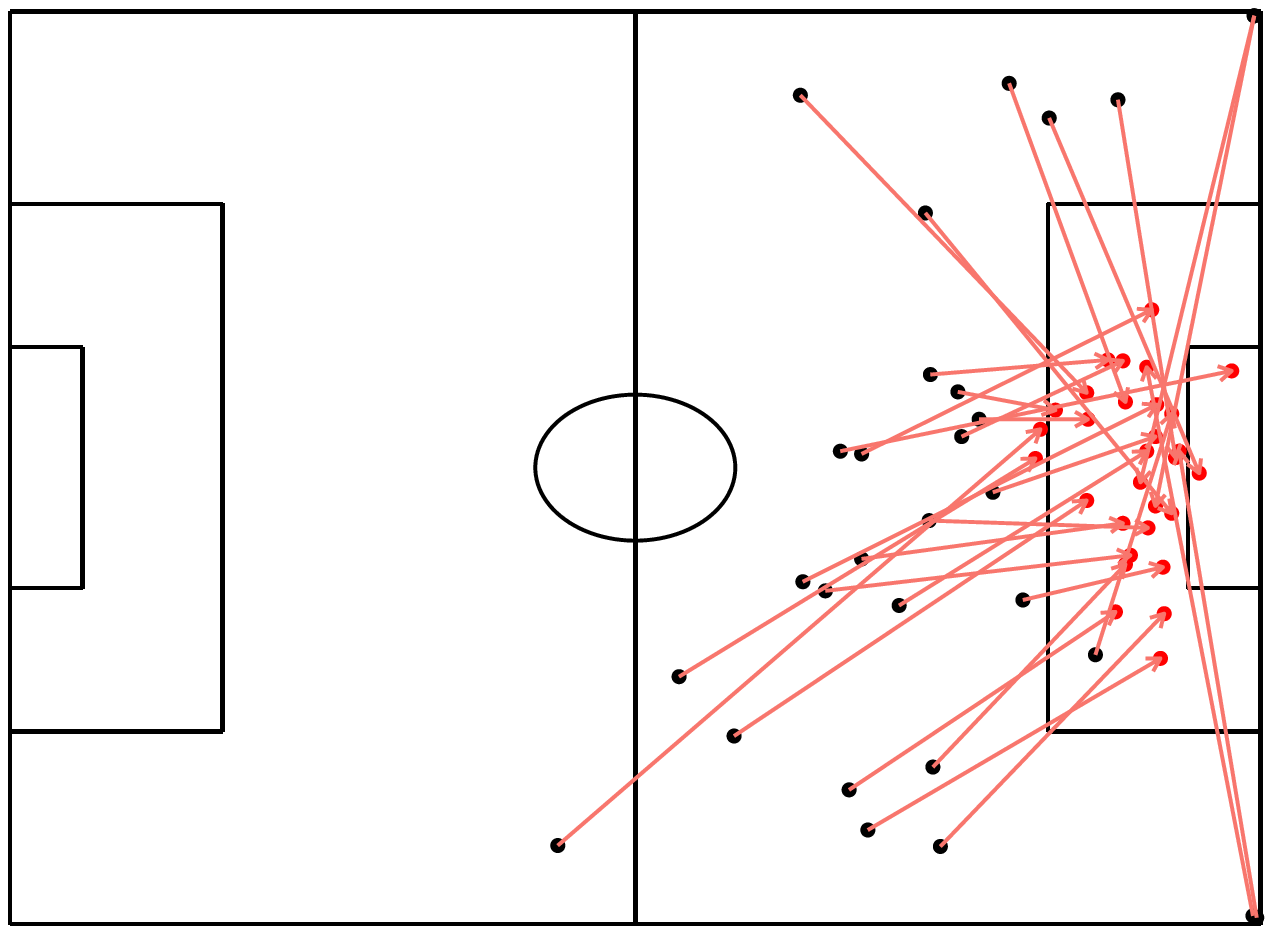}}
\subfigure[Luis Suarez]{\includegraphics[clip=true, trim=1cm 2.5cm 1cm 3cm,width=4cm]{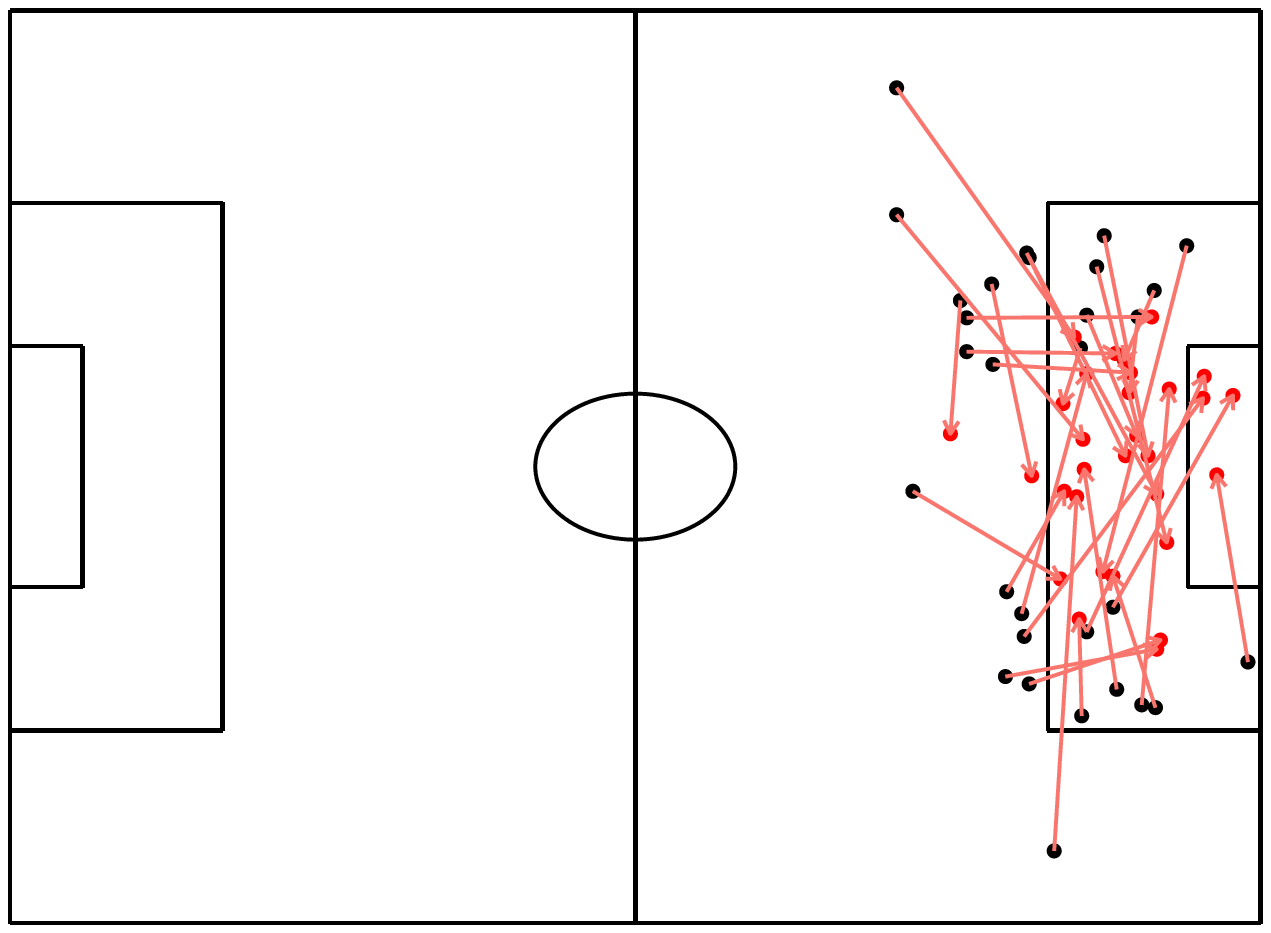}}
\vspace{-5mm}
\caption{The top 30 passes with the highest QPass value. QPass is able to distinguish the specific roles of players in the attack: playmaking or finishing.}
\label{fig:good_passes}
\end{figure}

We take a closer look at the passes the two types of attackers tend to make. Figure~\ref{fig:good_passes} presents the top 30 passes with the highest QPass values in case of Messi and Suarez. Messi helps his team by making passes from outside of the box into the opponent's box---a premier spot to shot on target. On the other hand, 
the passes of Suarez start already inside the box. These are lateral passes from a good part of the field to an even better one.



\subsection{Lose the ball to win the game}
\begin{figure}[tb]
\centering
\includegraphics[width=5.5cm]{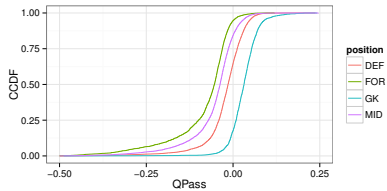}
\vspace{-5mm}
\caption{Team can be better off by losing the ball.}
\label{fig:losing_ball}
\end{figure}

It is part of the game to make errors, \ie, to lose the ball now and then. In fact, the constant change of ball possession is one of the main reasons why soccer is such a dynamic game. Losing the ball is considered to be a bad thing, however, as our study reveals, a team can be better off with a lost ball in some circumstances. The methodology of QPass is able to identify unsuccessful passes that were useful, \ie, increased the field value of a team. To justify this, we present the cumulative distribution of the QPass values of unsuccessful passes in Figure~\ref{fig:losing_ball}. Almost 50 percent of the failed passes of defenders are actually helping their team despite the fact that the opponent will possess the ball afterwards. Similarly, goalkeepers are making good decisions by risking a pass that may be unsuccessful in 80 percent of the time. In some particular cases, even attackers' ``mistakes'' are beneficial after all.

\begin{figure}[tb]
\centering
\subfigure[Gerard Pique]{\includegraphics[clip=true, trim=1cm 2.5cm 1cm 3cm,width=4cm]{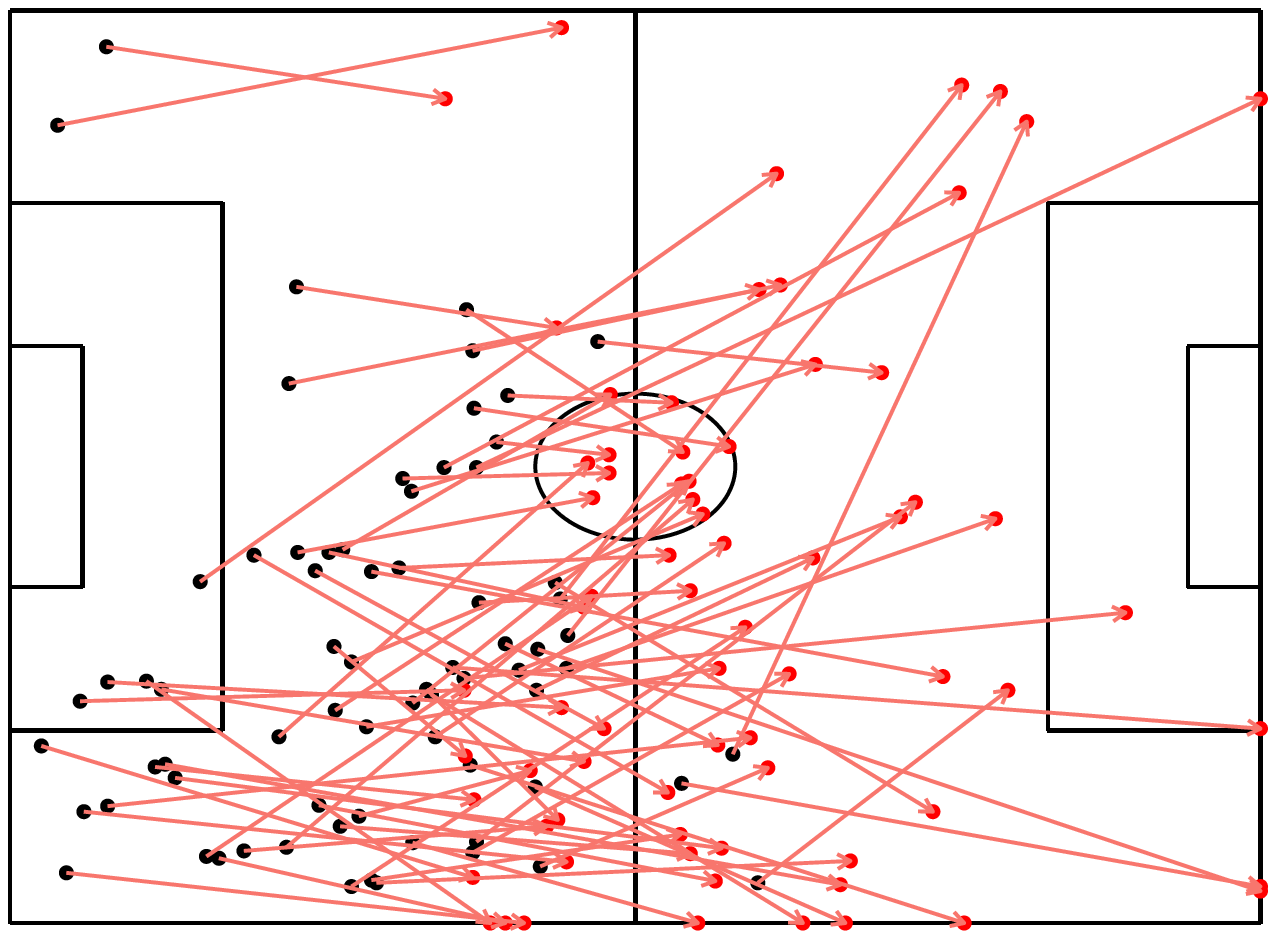}}
\subfigure[Lionel Messi]{\includegraphics[clip=true, trim=1cm 2.5cm 1cm 3cm,width=4cm]{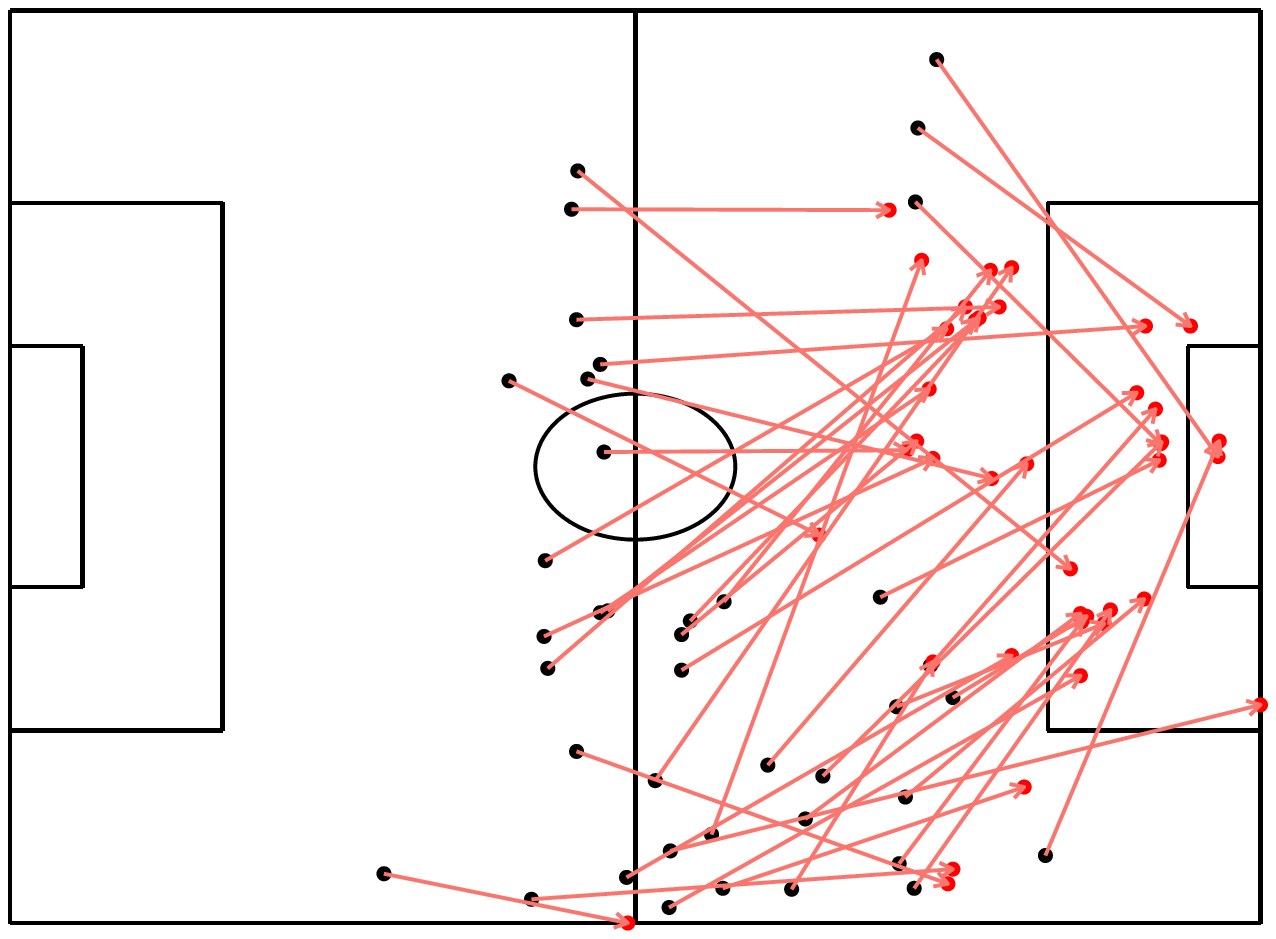}}
\vspace{-5mm}
\caption{Lose the ball to win: unsuccessful passes that lead to increased field values.}
\label{fig:lost_balls_beneficial}
\end{figure}

To give an idea on the nature of these scenarios, we present the unsuccessful passes of Pique and Messi of FC Barcelona that result in elevated field values (Figure~\ref{fig:lost_balls_beneficial}). In case of Pique, these passes are either clearances to avoid conceding shots or crosses to the wings to start building up an attack. Messi on the other hand tries to pass the ball into the attacking third (or even into the box of the opponent) to prepare immediate scoring opportunities.


\section{Conclusion}
We introduced a novel methodology called QPass to evaluate the intrinsic value of passes, \ie, how they contribute to having more scoring opportunities than the opponent. Through the analysis of an entire season, we rank the players based on their passing capabilities using QPass, and identify players best suitable for specific styles of play. Our counterintuitive finding reveals that making unsuccessful passes may lead to better chance to win a game. Applying QPass in practice can help player and opponent scouting by revealing the merit of the players' passing game.



%
\bibliographystyle{abbrv}
\small{
\vspace{1mm}
\bibliography{sigproc}  
}
%
%
\end{document}